**OPEN**

# Carbon Emission Prediction in China Considering New Quality Productive Forces Using a Deep & Corss Learning Modeling Framework

Haijin Xie[1,2], Gongquan Zhang[3,4] ✉

New quality productive forces (NQPF), digital economy advancement, and artificial intelligence (AI) technologies are becoming crucial for promoting sustainable urban development. This study proposes a Multi-head Attention Deep & Cross Network (MADCN) framework, combining feature interaction modeling and attention mechanisms, to predict urban carbon emissions and investigate the impacts of technological factors. The framework incorporates an interpretable learning phase using SHapley Additive exPlanations (SHAP) to assess the contributions of different features. A panel dataset covering 275 Chinese cities is utilized to test the MADCN model. Experimental results demonstrate that the MADCN model achieves superior predictive performance compared to traditional machine learning and deep learning baselines, with a Mean Squared Error (MSE) of 406,151.063, a Mean Absolute Error (MAE) of 612.304, and an R-squared ($R^2$) value of 0.991 on the test set. SHAP analysis highlights that population, city size, urbanization rate, and GDP are among the most influential factors on carbon emissions, while NQPF, digital economy index, and AI technology level also show meaningful but relatively moderate effects. Advancing NQPF, strengthening the digital economy, and accelerating AI technology development can significantly contribute to reducing urban carbon emissions. Policymakers should prioritize integrating technological innovation into carbon reduction strategies, particularly by promoting intelligent infrastructure and enhancing digitalization across sectors, to effectively achieve dual-carbon goals.

## 1. Introduction

With the intensifying challenges of climate change, energy consumption, and carbon emissions have become central issues in global environmental governance. The energy sector remains the dominant contributor to greenhouse gas emissions, accounting for over 70% of global $CO_2$ emissions [1]. In response, countries worldwide have implemented a wide range of carbon reduction policies, including renewable energy promotion, carbon trading markets, and regulatory caps on emissions [2, 3]. In recent years, emerging forces such as the digital economy and artificial intelligence (AI) have demonstrated considerable potential in promoting low-carbon development by optimizing energy structures and improving resource efficiency [4, 5]. However, a newly conceptualized variable—new quality productive forces (NQPF)—has become a transformative driver of high-quality and sustainable economic development in Chinese policy discourse. NQPF, while unique to China's policy context, aligns with and extends established theories: it integrates technological spillovers (endogenous growth), industrial upgrading (structural transformation), and institutional transformation into a policy - driven ecosystem. Distinct from isolated Western concepts, it leverages China's dual-carbon strategy to channel technological progress into decarbonization - aligned indus-

[1]Department of Philosophy, National Academy of Governance, Beijing, 100091, China. [2]Confucius Research Institute, Renmin University of China, Beijing, 100872, China. [3]School of Traffic and Transportation Engineering, Central South University, Changsha, 410075, China.[4]Harvard University, Boston, MA 02138, USA. ✉email: 214201028@csu.edu.cn.

-trial evolution, reconfiguring labor/capital via institutional tools like 14th Five - Year Plan policies. Rooted in technological innovation, ecological efficiency, and labor-capital recombination, NQPF is expected to play a critical role in carbon mitigation [6, 7]. Yet, the mechanisms by which NQPF influences carbon emissions remain largely unexplored in academic literature, and its potential as a policy-relevant variable in climate governance is still theoretically and empirically underdefined.

Previous studies examining the impact of emerging variables such as digitalization and AI on carbon emissions have predominantly relied on traditional statistical models, including ordinary least squares (OLS), fixed effects (FE) panel regressions, and generalized method of moments (GMM) [8]. These methods offer clear interpretability and well-established inference frameworks based on predefined assumptions about linearity, normality, and independence. While statistical models provide insights into average marginal effects and causal inference, they are inherently limited in capturing complex nonlinear relationships and high-order feature interactions. To address these limitations, recent studies have introduced machine learning (ML) techniques, such as random forests, support vector machines (SVM), and gradient boosting decision trees (GBDT), into carbon emission forecasting [9]. ML models enhance prediction accuracy by leveraging data-driven learning without strict parametric assumptions. However, their performance is still constrained by feature engineering limitations, lack of temporal adaptability, and challenges in capturing higher-level semantic interactions. These shortcomings have led to the emergence of deep learning models as a promising alternative.

Deep learning models, such as multilayer perceptrons (MLP), convolutional neural networks (CNN), and recurrent neural networks (RNN), are capable of modeling complex, nonlinear, and hierarchical patterns in high-dimensional data [10, 11, 12]. These models learn latent representations through multiple hidden layers, enabling them to outperform traditional approaches in many forecasting and classification tasks, including environmental modeling. Their advantages include strong generalization capabilities, high robustness to noise, and the ability to learn from unstructured data [5, 13]. However, two critical limitations persist. First, most deep learning models treat input variables as isolated entities and often neglect interactions among features, which can significantly influence output behavior in environmental systems. Second, deep learning models are frequently criticized for their lack of interpretability, making it difficult to understand how specific variables contribute to predictions, especially problematic in public policy and environmental decision-making. Given the focus on carbon emission prediction stems from critical policy and theoretical needs, China's dual-carbon goals demand city-specific emission forecasts to design targeted decarbonization strategies, but existing models fail to capture interactions among emerging factors and lack interpretability.

To overcome the aforementioned limitations, this study proposes a Deep & Cross Learning framework for carbon emission modeling. Specifically, a Multi-head Attention Deep & Cross Network (MADCN) is proposed to capture the individual and interactive effects of key variables, such as new quality productive forces, digital economy development, and AI penetration, on regional carbon emissions. The architecture integrates cross-feature transformation and attention mechanisms to model feature interactions explicitly. Furthermore, we apply SHapley Additive exPlanations (SHAP) to quantify each variable's marginal and interaction effects, thereby enhancing model transparency. Finally, a sensitivity analysis is conducted to assess the dynamic influence of each driver on carbon emissions under future development scenarios. The main contributions of this study are threefold: (1) an extensible modeling framework that incorporates variable feature values and their interaction effect; (2) a MADCN model as the novel architecture tailored for carbon emission prediction; and (3) quantifiable interpretation of variable impacts using SHAP and sensitivity analysis, offering actionable insights for low-carbon policy formulation.

The rest of this paper is organized as follows: Section 2 reviews the related literature on new quality productive forces, carbon emissions, and AI-based modeling methods. Section 3 describes the proposed methodology, including the MADCN model structure and the SHAP interpretability approach. Section 4 introduces the data sources, variable construction, and preprocessing steps. The results and discussion are presented in Section 5, highlighting both the prediction performance and interpretative findings. Section 6 concludes this study, discusses its theoretical and policy implications, and outlines directions for future research.

## 2. Literature review

This section presents a comprehensive review from two perspectives: the influencing factors of carbon emissions and the development of carbon emission prediction models.

### 2.1 Driving Factors of Carbon Emissions

Previous studies have examined the driving forces behind carbon emissions. Traditionally, carbon emissions have been primarily attributed to factors such as economic growth, population size, industrial structure, energy consumption patterns, urbanization, and technological advancement [14, 15, 16]. According to the Environmental Kuznets Curve (EKC) hypothesis, carbon emissions tend to rise with economic development up to a certain threshold, after which they begin to decline as cleaner technologies and environmental policies are adopted [17]. The scale effect, structural effect, and technology effect are frequently used to explain the mechanism by which economic activity influences carbon output. The scale effect emphasizes that economic expansion usually leads to increased energy consumption and emissions. The structural effect suggests that the shift from heavy to light industry or service sectors can reduce emissions intensity [18]. The technology effect highlights the role of innovation in improving energy efficiency and promoting clean energy. However, these conventional factors alone cannot fully explain the complex and evolving patterns of carbon emissions observed under digital transformation.

In recent years, the concept of the digital economy has been increasingly recognized as a key factor in shaping carbon emission dynamics. The digital economy, characterized by internet infrastructure, cloud computing, big data, and e-commerce, alters traditional energy consumption and production modes [19]. It contributes to emissions reduction by optimizing supply chain management, enhancing operational efficiency, and supporting remote work and intelligent logistics [20]. At the same time, the infrastructure supporting digitalization, such as data centers and ICT equipment, may also lead to increased energy demand [21]. Therefore, the digital economy influences carbon emissions through a dual mechanism: facilitating decarbonization through efficiency gains while potentially inducing new sources of energy consumption. The net effect depends on factors such as technological maturity, energy mix, and the scale of digital adoption [22]. Besides digitalization, the rapid rise of artificial intelligence (AI) has attracted scholarly attention as another transformative factor affecting carbon emissions. AI technologies, including machine learning, computer vision, and natural language processing, enhance automation, energy demand forecasting, and system optimization in various sectors such as transportation, manufacturing, and energy [4, 11, 23]. AI supports low-carbon transitions by enabling precision energy management, intelligent environmental monitoring, and real-time optimization of resource allocation. Moreover, AI and the digital economy exhibit a strong degree of technological convergence. Digital platforms serve as enablers for AI deployment, while AI enhances the intelligence and adaptability of digital systems [5, 24]. This mutual reinforcement implies that the joint effect of AI and digitalization on carbon emissions may differ significantly from their isolated impacts, suggesting a potential non-linear or interaction-based influence pathway.

Recently, the notion of new quality productive forces (NQPF) has emerged as a strategic concept within China's national development framework. NQPF represents a leap in productive capabilities characterized by advanced technologies, green development principles, and the restructuring of labor-capital relations [6, 25]. Preliminary studies indicate that NQPF may promote carbon mitigation by accelerating technological innovation, fostering high-efficiency industries, and reshaping economic growth models [26, 27]. Compared to digitalization and AI, NQPF is a more integrative construct that encompasses multiple dimensions of modernization and sustainability. Some researchers argue that NQPF synthesizes the enabling effects of digitalization, AI, and green transformation, thereby exerting a compounded influence on emission outcomes [28, 29]. Substantively, the digital economy provides foundational infrastructure (e.g., data platforms, smart grids) enabling large-scale AI-driven optimizations, while AI enhances the efficiency of digital systems (e.g., reducing energy waste in data centers or optimizing supply chains). This synergy can generate an "amplification effect": for example, digital platforms facilitate real-time energy monitoring, and AI algorithms process this data to dynamically adjust industrial energy use, yielding greater emissions reductions than either technology alone. Conversely, offsetting effects may occur if the digital economy expands energy-intensive infrastructure (e.g., data centers) faster than AI can optimize their efficiency, resulting in net emission increases. These dynamics align with the logic of NQPF, which emphasizes the fusion of technology, industry, and institutions—non-linear interactions capture how these fused elements jointly shape decarbonization, rather than acting in isolation.

However, the empirical exploration of how NQPF affects carbon emissions, particularly in conjunction with or beyond its component variables, remains limited [30]. The potential interactive effects between NQPF, AI, and the digital economy are underexplored, and the exact mechanisms through which these factors influence carbon dynamics are not yet fully clarified. There is a pressing need to explore whether NQPF not only independently affects carbon emissions but also alters the relationship between digital and intelligent technologies and environmental outcomes. Furthermore, the dynamic evolution of NQPF and its embedded interaction with urbanization levels, industrial upgrading, and policy innovations such as green finance and emission trading systems presents additional layers of complexity that require sophisticated analytical frameworks.

### 2.2 Carbon Emission Prediction Models

The development of carbon emission prediction models has followed three major methodological trajectories: statistical models, traditional machine learning models, and deep learning models.

Statistical models are among the earliest tools used to forecast carbon emissions. These models, grounded in econometric theory, typically include linear regression, autoregressive integrated moving average (ARIMA), vector autoregression (VAR), cointegration models, and panel data regressions [8, 31, 32]. These methods rely on predefined functional forms and theoretical assumptions regarding stationarity, linearity, and error distribution. Statistical models are advantageous for their interpretability and causal inference capability, offering insights into the marginal effects of predictors on emissions [33]. However, their major limitation lies in their inability to capture nonlinear dependencies, dynamic feature interactions, and time-varying structures that are increasingly common in carbon emission data. As a result, researchers have gradually introduced traditional machine learning (ML) techniques [9, 34, 35]. Traditional machine learning models overcome some of the constraints of statistical models by allowing more flexible and data-driven learning frameworks. Common ML models used in emission prediction include decision trees (DT), random forests (RF), support vector machines (SVM), k-nearest neighbors (KNN), gradient boosting decision trees (GBDT), adaptive boosting (AdaBoost), and extreme gradient boosting (XGBoost) [36, 37, 38]. These models are well-suited for high-dimensional and non-linear datasets, providing improved prediction accuracy and robustness. Nonetheless, traditional ML models depend heavily on manual feature engineering, lack long-term temporal learning capabilities, and often struggle with interpretability. Their performance may also deteriorate in the presence of complex multi-variable interactions or noisy input data. These limitations have paved the way for the integration of deep learning (DL) models into the carbon forecasting field.

Deep learning models, derived from neural network architectures, have become prominent in carbon emission prediction due to their capacity to model nonlinear, hierarchical, and sequential relationships [4, 5, 39]. These models are especially powerful in recognizing hidden patterns and extracting hierarchical representations from structured and unstructured datasets, enabling them to adapt to complex and dynamic carbon emission systems. Typical DL models include multilayer perceptrons (MLP), convolutional neural networks (CNN), recurrent neural networks (RNN), long short-term memory networks (LSTM), gated recurrent units (GRU), autoencoders, and transformer-based architectures such as BERT and Vision Transformers (ViTs) [11, 12, 40, 41]. These models automatically extract complex features from raw data, adapt to temporal dynamics, and process spatial-temporal patterns, making them highly suitable for forecasting tasks involving large-scale and time-series emission data. For instance, LSTM and GRU are particularly effective in handling long-term dependencies in carbon emission time series data, where past events may have delayed but significant effects on future outcomes [42]. CNNs and ViTs, on the other hand, are advantageous in capturing spatial correlations across geographical regions or industrial zones [43]. Moreover, transformer-based models equipped with attention mechanisms allow the selective focusing on relevant temporal or spatial features, which is especially valuable in scenarios involving heterogeneous environmental data [44]. These models are also highly scalable and can incorporate multi-source and multi-frequency datasets, such as satellite imagery, industrial output reports, and policy indices, into a unified learning pipeline. Such integration is critical for advancing high-resolution carbon accounting and real-time emission monitoring.

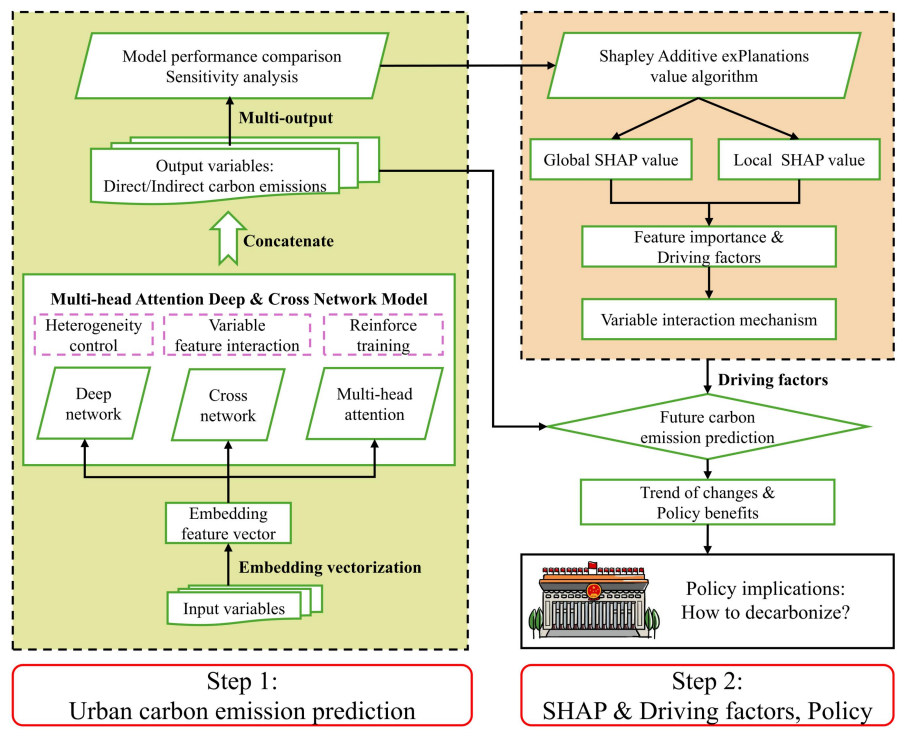

**Figure. 1.** Deep & Cross modeling framework

However, conventional DL models treat each feature independently and often neglect the interactions among features, which may lead to suboptimal performance in understanding composite drivers of emissions. Such interactions, for example between economic growth and energy structure, or between digital infrastructure and urbanization, are often crucial in shaping emissions but are implicitly overlooked. DL models are normally regarded as "black boxes" due to their lack of interpretability, which limits their applicability in policy-oriented decision-making and hinders stakeholder trust [45]. Without a clear understanding of how inputs influence outputs, especially when predicting sensitive policy outcomes, the adoption of such models in regulatory and strategic environmental planning remains constrained. Enhancing interpretability and modeling feature interactions are key areas of improvement for the next generation of carbon emission forecasting tools.

Despite extensive research on carbon emission drivers and predictive modeling, several key limitations remain. 1) The impact of new quality productive forces (NQPF) on carbon emissions has not been clearly established, and the interactive effects between NQPF, the digital economy, and artificial intelligence (AI) remain largely unexamined. These interactions may produce synergistic or non-linear effects that are crucial for understanding carbon dynamics in the context of emerging technological paradigms. 2) Most existing carbon emission forecasting models lack transferability and scalability, as they are typically designed for specific datasets, regions, or temporal scopes, limiting their applicability in broader or evolving contexts. 3) Although deep learning models have demonstrated superior performance in handling complex, high-dimensional data, they often fail to consider feature interactions, and their lack of interpretability restricts their use in policy-making, where understanding variable contributions and causal relationships is critical.

## 3. Methodology

This study develops a Deep & Cross modeling framework for predicting and analyzing urban carbon emissions, as shown in **Fig. 1**.

In the first step, a Multi-head Attention Deep & Cross Network (MADCN) model is constructed to predict urban carbon emissions. MADCN utilizes a deep network to import input variables to build the relationship between factors and urban carbon emissions. Then, it designs a cross network to integrate the interactive effects of various variables. The multi-head attention mechanism is employed to enhance the model training speed and stability. The second step employs the SHapley Additive exPlanations (SHAP) method to determine the influence of various variables on urban carbon emissions. SHAP value identifies the driving factors of transportation carbon emissions and their interactive mechanisms.

### 3.1 Attention random deep & cross network

This paper proposes a novel Multi-head Attention Deep & Cross Network (MADCN) model to mitigate the impacts of heterogeneity on the model prediction accuracy and to consider the effects of feature interactions on carbon emissions. MADCN comprises several components: embedding layer, deep network layer, cross network layer, multi-head attention layer, fusion layer, and output layer, as illustrated in **Fig. 2**.

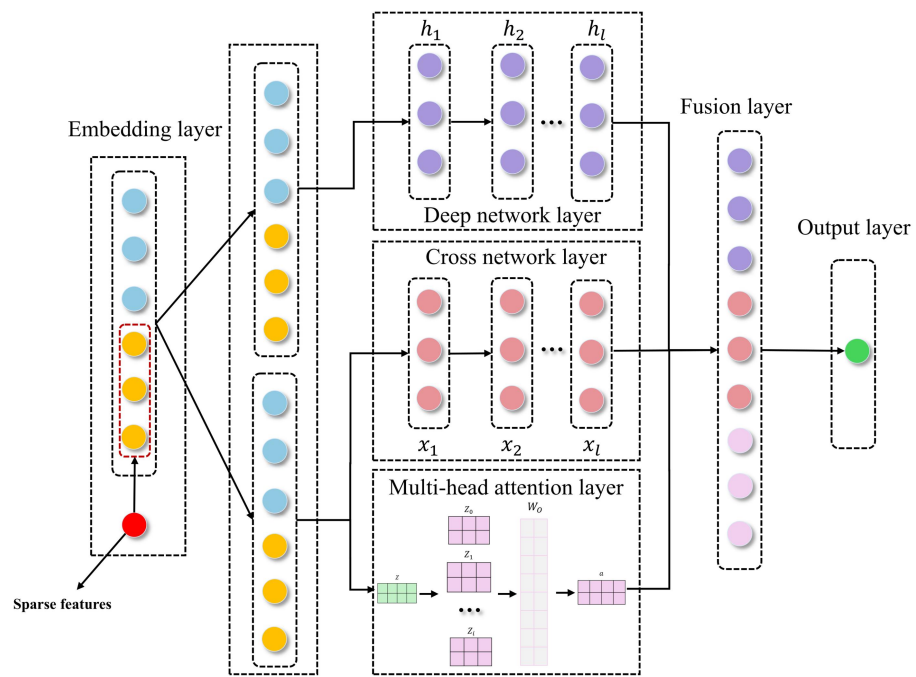

**Figure. 2.** Architecture of Multi-head Attention Random Deep & Cross Network

Dense features include transportation energy consumption, transportation development level, etc. Sparse features consist of city ID and year. MADCN employs the embedding vectorization method to transform sparse features into dense vectors stacked with dense features and separately input into neural network layers. The feature vectors processed by the embedding layer are directly input into the cross network, establishing a nonlinear relationship between feature interactions and carbon emissions. Then, it is input into the deep network layer to establish the nonlinear relationships between variables and carbon emissions. The multi-head attention layer captures the multi-level relationships between variables, enhancing the model's robustness and stability. The fusion layer integrates outputs from the above network structures to predictions for transportation industries' direct and indirect carbon emissions.

### 3.2.1 Embedding layer & Random parameter

The embedding vectorization method transforms sparse features of discrete variables into low-dimensional dense vectors through embedding matrices. For continuous variables, they are typically inputted directly into the model and subjected to standardization or normalization. Assuming the embedding dimension is $d$, for a discrete variable $x_i$, its embedding vector $x_{v,i}^T$ is expressed as:

$$x_{v,i}^T = W_i e_i, x_{v,i}^T \in \mathbb{R}^d \tag{1}$$

where $W_i$ is the embedding matrix for the discrete variable $x_i$.

For dense features of continuous variables $x_j$, the normalization is given by:

$$x_i^{'} = \frac{x_j - \mu_j}{\sigma_j} \qquad (2)$$

where $\mu_j$ and $\sigma_j$ represent the mean and standard deviation of $x_j$, respectively.

The model combines dense and sparse features into a single feature vector for input into the network layers. Suppose there are $m$ continuous features and $n$ sparse features, with the dense feature vector being $X_c \in \mathbb{R}^m$ and the embedded sparse feature vector $X_v \in \mathbb{R}^m$:

$$Z = [X_c, X_v] = [x_{dense}^T, x_{v,1}^T, x_{v,2}^T, x_{v,3}^T, ..., x_{v,i}^T] \qquad (3)$$

where $X_c$ represents the vector of all dense features, and $X_v$ is the concatenation of all embedding vectors. $x_{dense}^T$ represents the dense feature vector, and $x_{v,i}^T$ contains the embedding vectors of $i$ sparse feature.

For the cross network, the input is $Z$. For the deep network, the random parameter generated by a Gaussian noise generator is introduced:

$$R \sim \mathcal{N}\left(x \middle| \mu, \sigma^2\right) = \frac{1}{\sqrt{2\pi\sigma^2}} e^{-\frac{(x-\mu)^2}{2\sigma^2}} \qquad (4)$$

where $x$ is the variable, $\mu$ represents the mean value, $\sigma^2$ is the variance, and $\sigma$ is the standard deviation. Therefore, the input to the deep network layer:

$$Z^{'} = Z + R \qquad (5)$$

where $R$ is the random parameter.

### 3.2.2 Deep & cross network

The cross network layer is designed to capture high-order feature interactions. Each cross network layer forms a new feature representation by element-wise multiplying the input features with the output from the previous layer (see **Fig. 3**). Through multiple layers of cross operations, the model captures the high-order nonlinear interactions between features. The formula for this process is:

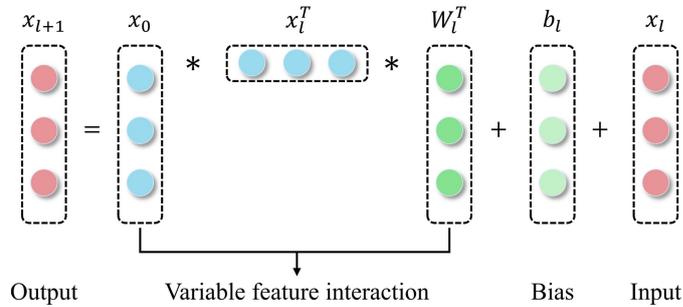

$$x_{l+1} \qquad x_0 \qquad * \qquad x_l^T \qquad * \qquad W_l^T \qquad b_l \qquad x_l$$

Output     Variable feature interaction     Bias     Input

**Figure. 3.** Principle of cross network layer

$$x_{l+1} = x_0 \cdot \left(W_l^T x_l^T + b_l\right) + x_l \qquad (6)$$

where $x_0$ is the original input feature vector $Z$, $x_l$ is the output feature from the $l$-th layer, and $W_l^T$ and $b_l$ are the weight and bias of the $l$-th layer, respectively.

The deep network layer consists of a multi-layer perceptron (MLP) made up of several fully connected neural networks, each followed by a nonlinear activation function. This setup allows for learning complex nonlinear representations of input features. Through multiple layers of nonlinear transformations, the model captures deep patterns and complex relationships in the input features.

$$h_l = f(W_l h_{l-1} + b_l) \qquad (7)$$

where $h_o$ is the input feature vector $Z^{'}$, $W_l$ and $b_l$ are the weight and bias of the $l$-th layer, and $f(\cdot)$ represents the activation function.

### 3.2.3 Multi-head attention mechanism

The multi-head attention mechanism (see **Fig. 4**) is employed to capture significant relationships between input features, enhancing the model's ability to select important features through interactions in different subspaces. By incorporating attention mechanisms, the model focuses on crucial feature interactions, thereby improving the quality of feature representation and overall model performance.

$$Q = ZW_Q, K = ZW_K, V = ZW_K \tag{8}$$

where $Q$, $K$, and $V$ are the query, key, and value vectors, respectively. $W_Q$, $W_K$, $W_K$ are trainable weight matrices. The attention function is defined as:

$$\text{Attention}(Q, K, V) = \text{softmax}\left(\frac{QK^T}{\sqrt{d_k}}\right)V \tag{9}$$

where $d_k$ represents the dimension of the key vector. The multi-head attention is formulated as:

$$\text{MultiHead}(Q, K, V) = \text{Concat}(head_1, head_2, ..., head_i)W_O \tag{10}$$

$$head_i = \text{Attention}(Q_i, K_i, V_i) \tag{11}$$

where $W_O$ is the linear transformation matrix for the output. $Q_i$, $K_i$, and $V_i$ are subsets of the query, key, and value vectors processed by the $i$-th attention head, respectively.

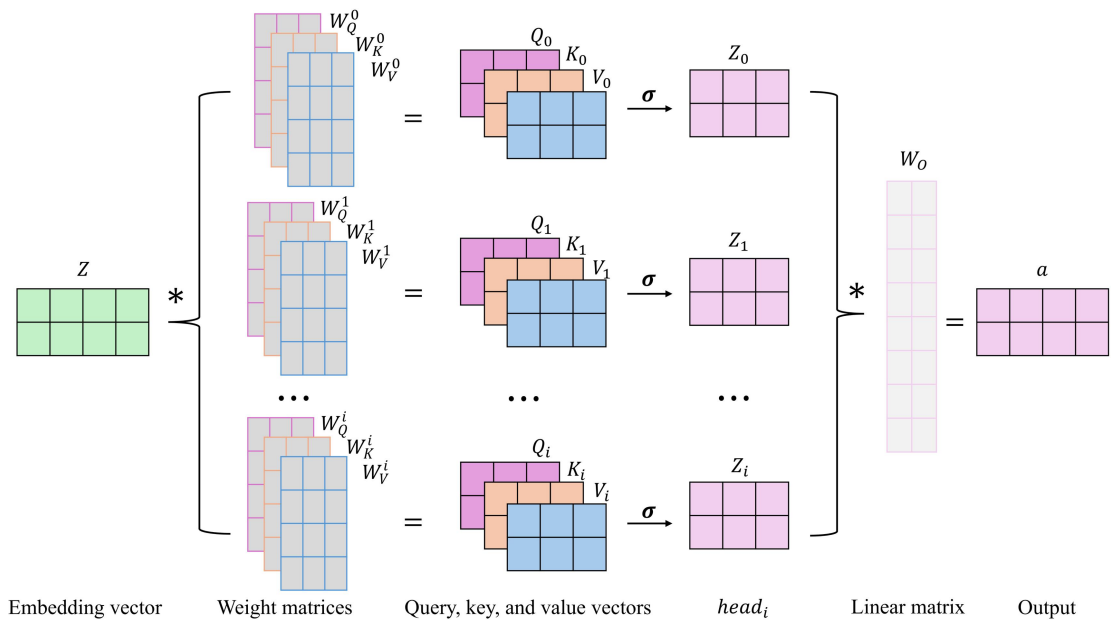

**Figure. 4.** Principle of multi-head attention layer

### 3.2.4 Model learning process

In the MADCN model, outputs from the cross network layer, deep network layer, and multi-head attention layer are concatenated and processed through a fully connected neural network for prediction.

$$Z_{final} = [x_L, h_L, a] \tag{12}$$

where $x_L$ is the output from the last cross network layer, $h_L$ is the output from the last deep network layer, and $a$ is the output from the multi-head attention layer.

$$\hat{y} = \sigma(W_o Z_{final} + b_o) \tag{13}$$

where $W_o$ and $b_o$ are the weights and biases of the output layer, and $\sigma$ is an activation function.

The MADCN model is trained through backpropagation and gradient descent algorithms for concurrent objective prediction. The loss function is mean squared error (MSE), minimized using the Adam optimizer.

$$\mathcal{L} = \frac{1}{N}\sum_{i=1}^{N}\left[(y_{1,i} - \hat{y}_{1,i})^2 + (y_{2,i} - \hat{y}_{2,i})^2\right] \tag{14}$$

where $y_{1,i}$ and $y_{2,i}$ represent the true values of the $i$-th sample, $\hat{y}_{1,i}$ and $\hat{y}_{2,i}$ represent the predicted values of the $i$-th sample, and $N$ is the number of samples. The gradient descent optimizer updates parameters according to the formula:

$$\theta \leftarrow \theta - \eta\frac{\partial\mathcal{L}}{\partial\theta} \tag{15}$$

where $\theta$ represents the model parameters such as weights and biases, and $\eta$ is the learning rate.

In MADCN, the multi-head attention mechanism and cross network function synergistically to capture the complexity of carbon emission drivers. The multi-head attention mechanism dynamically assigns weights to variables based on their contextual relevance. For instance, it prioritizes AI and digital infrastructure in cities with mature industrial systems, while emphasizing traditional factors like population density in less developed regions. This ensures that context-sensitive drivers (e.g., NQPF's role in moderating tech-industry interactions) are effectively captured by focusing on "which variables matter under specific conditions."

The cross network, by contrast, explicitly models high-order feature interactions critical to structural decarbonization. While the deep feedforward network captures general nonlinear patterns, the cross network directly quantifies multiplicative relationships—such as the amplification of AI's energy optimization effects by digital economy infrastructure, or the offsetting of digital expansion's emission intensity by NQPF. This ensures that synergies and offsets between emerging variables are explicitly learned, rather than being implicit in the model. Together, the attention mechanism identifies contextually relevant variables, and the cross network unpacks their joint effects, enabling MADCN to simultaneously capture both individual and combined impacts of diverse drivers, a capability central to predictive and interpretive power.

### 3.3 Model interpretation

After employing hyperparameter tuning and cross-validation techniques, the optimal MADCN model and parameter combination are obtained. This optimal prediction model then serves as the foundational model for the Shapley Additive exPlanations (SHAP) value algorithm, quantifying each factor feature's contribution values.

SHAP is an explanatory method that measures the importance of features based on the characteristics of the machine learning model itself. SHAP values calculate the marginal contribution of each feature in a sample and possess characteristics such as consistency (the importance of a feature does not decrease when the model becomes more dependent on it), additivity, symmetry, nullity (a SHAP value of zero indicates no contribution), and linearity (marginal contributions are linearly additive). Compared to feature importance and permutation importance, SHAP allows for consistent analysis of individual sample feature importance.

Suppose $f(x)$ is a well-trained classifier, and $g(x)$ is an additive model used to interpret model evaluation results. In this case, the SHAP value for each sample is represented as:

$$f(x_m) = g(x_m) = \varphi_0 + \sum_{i=1}^{n} \varphi_i^m \qquad (16)$$

$$\varphi_i^m = \sum_{S} \frac{|S|! \, (n - |S| - 1)!}{n!} [f_x(S \cup \{x_i\}) - f_x(S)] \qquad (17)$$

where $n$ represents the number of features, $m$ represents the $m$-th sample in the test set, and $x_i$ is the $i$-th feature in the sample. $\varphi_0$ is the baseline prediction value, which is the average of all sample SHAP values. $\varphi_i^m$ is the SHAP value for the feature $x_i$, quantifying its impact on the model output. The set $S$ represents all feature subsets that exclude feature $x_i$; $f_x(S \cup \{x_i\})$ and $f_x(S)$ are the predicted values with and without the feature $x_i$, respectively, referring in this article to the direct or indirect carbon emissions in the transportation industry. The absolute value of a feature's SHAP value indicates the significance of its impact on the prediction.

## 4. Data source and preparation

This section briefly introduces the data sources and index selection process used to validate the proposed carbon emission prediction framework.

### 4.1 Data sources

The dataset used in this study is compiled from a wide range of Chinese statistical yearbooks and official online platforms, ensuring accuracy, consistency, and credibility. Core data were obtained from the China Statistical Yearbook (2000–2021), China Urban Construction Statistical Yearbook, China Regional Economic Statistical Yearbook, China Environmental Statistical Yearbook, China Transportation Yearbook, China Science and Technology Statistical Yearbook, and the China Digital Economy Development Report.

Indicators related to new quality productive forces (NQPF) were primarily constructed based on composite proxies, integrating high-tech industry output, R&D investment intensity, labor productivity in strategic emerging sectors, and innovation inputs at the city level. These data were retrieved and aggregated from the China Science and Technology Statistical Yearbook, China Statistical Yearbook on High Technology Industry, and the municipal statistical yearbooks of selected cities. Indicators of the digital economy were collected from the China Digital Economy Development Report and the Digital Finance Inclusion Index Database released by Peking University, as well as broadband coverage, mobile internet usage, and smart infrastructure investment data extracted from municipal government reports and telecommunications authority bulletins. dditional variables describing the development of AI were collected from the China Artificial Intelligence Industry Development Report, the AI Open Index Database, and the official websites of major city statistical bureaus. These include AI patent counts, AI industry value added, and AI-related policy initiatives at the city level.

To address issues of missing values or inconsistent year coverage in certain sources, a supplemental open-source dataset—Carbon Emission Prediction for 275 Chinese Cities (2000–2021)—was utilized. This dataset, publicly available at https://github.com/Sekiro23268/Carbon-Emission-Prediction, provides harmonized and quality-assured carbon emission and urban indicator data. After cleaning and integration, the final dataset comprises over 720,000 monthly-level observations across 275 prefecture-level cities in China, covering the years 2009-2021.

### 4.2 Index selection

To comprehensively explore the impact of new productive factors on carbon emissions, this study constructs a multidimensional indicator framework centered on four core dimensions: NQPF, digital economy development, AI advancement, and urban carbon emissions. These dimensions reflect the evolving structure of high-quality development and its environmental implications.

NQPF indicators include high-tech industry output share, R&D intensity, and labor productivity in innovation-intensive industries. In detail, chosen as a proxy for NQPF's technological dimension, robot density reflects policy-directed technological spillovers, aligning with endogenous growth theory. Employees in strategic emerging industries capture NQPF's industrial upgrading logic, distinct from generic productivity-focused upgrading by emphasizing decarbonization alignment. Digital economy indicators comprise of internet penetration rate, digital infrastructure investment, and digital financial inclusion. AI-related indicators include the number of AI patents, AI industry output, and AI policy support index.

In addition to these main variables, the model also incorporates a range of control variables that reflect fundamental city-level characteristics. These include population size, GDP per capita, urbanization rate, industrial structure (e.g., share of tertiary industry), green space ratio, public transportation coverage, infrastructure investment, and local environmental policy intensity.

The integration of emerging and traditional indicators allows the proposed model to effectively capture not only the direct impact of NQPF, digital economy, and AI on carbon emissions, but also their interaction with baseline urban development factors. This multidimensional design enables a more robust and interpretable analysis of urban carbon emission dynamics in the context of China's transition toward innovation-driven and low-carbon growth.

### 4.3 Input and output index

**Table 1** presents the selected indicators, where explanatory variables are used as model inputs to predict the target outcome, i.e., urban carbon emissions. The dataset includes a comprehensive set of indicators reflecting urban development, economic activity, industrial structure, and environmental sustainability across Chinese cities from 2009 to 2021. Variables such as population, GDP, city size, and city development level are treated as direct structural indices indicating the demographic and economic status of cities. Binary indicators, including low-carbon city, carbon peak city, and smart city, represent urban strategic policy implementations. These variables take values of 0 or 1, reflecting whether the corresponding policy has been adopted.

The urbanization rate, industry agglomeration level, AI technology level, and environmental pollution index are calculated based on ratios or composite indices. For example, industry agglomeration level is measured as the employment density per administrative area, and urbanization rate is computed as the proportion of urban population to total population. The environmental pollution index integrates multiple environmental quality measures using the golden triangle method.

The Digital Economy Index is constructed via entropy method (to weight individual indicators) and principal component analysis (to aggregate into core components), followed by standardization (z-score transformation) for comparability, with negative values indicating performance below the sample average. Similarly, the NQPF indicator reflects the comprehensive productivity of strategic sectors such as AI, robotics, telecommunications, and higher education. It is calculated through an integrated evaluation of inputs such as the number of AI-related companies, broadband subscriptions, and robot installation density. Carbon emissions represent total city-level emissions and are computed by aggregating emissions from various sectors, including transportation, construction, energy, agriculture, waste, and land use.

City ID, Year, and the above input indices serve as features for subsequent machine learning or statistical modeling frameworks used for heterogeneity analysis and carbon emission prediction. Insignificant features may be filtered out during modeling to improve generalization and interpretability.

## 5. Results and discussion

This section evaluates the effectiveness of the proposed Deep & Cross modeling framework for carbon emission prediction. The analysis proceeds in three stages. First, the dataset is divided into a training set and a testing set to validate the predictive capability of the model. Specifically, the complete dataset comprising N=41,449 observations is randomly partitioned, with 75% (N= 31,086) used for training and 25% (N=10,363) reserved for testing. Second, the predictive performance of the proposed MADCN model is systematically compared with several baseline approaches, including traditional statistical models, classical machine learning algorithms, and standard deep learning architectures. Finally, the interpretability of the MADCN framework is examined using SHAP, which quantifies the contribution of individual features and their interactions to the model output.

### 5.1 Evaluation metric

To evaluate the prediction performance of the MADCN model, this study employs three widely adopted metrics: Mean Squared Error (MSE), Mean Absolute Error (MAE), and the coefficient of determination ($R^2$).

MSE is defined as the average of the squared deviations between predicted values and actual observations. It is particularly sensitive to large errors due to the squaring operation, making it a suitable choice when large prediction deviations are especially undesirable. A smaller MSE value indicates higher model accuracy. The formula is as follows:

$$MSE = \frac{1}{n}\sum_{i=1}^{n}(y_i - \hat{y}_i)^2 \tag{18}$$

where $n$ represents the number of samples, $y_i$ is the observed value, and $\hat{y}_i$ is the predicted value.

**Table 1**
Factor and variable definitions

| Output index | Unit | Input index | Unit |
|---|---|---|---|
| Population | $10^4$ | Population | $10^4$ |
| GDP | *yuan* | GDP | *yuan* |
| Urbanization rate | % | Urban population | $10^4$ |
| | | Rural population | $10^4$ |
| City size | [1,7] | City size | [1,7] |
| City development level | [1,5] | City development level | [1,5] |
| Industry agglomeration level | | Number of employment | $10^4$ |
| | | Administrative area | $km^2$ |
| Environmental pollution index | | Air quality index | |
| | | Water quality index | |
| | | Pollution monitoring index | |
| Low-carbon city | {0,1} | Low-carbon city | {0,1} |
| Carbon peak city | {0,1} | Carbon peak city | {0,1} |
| Smart city | {0,1} | Smart city | {0,1} |
| New quality productive forces | | Employees in Strategic Emerging Industries | |
| | | Average Wage of Employees | $10^4$ *yuan* |
| | | General Higher Education Institutions | |
| | | Internet Broadband Subscribers | |
| | | Telecommunication Services | $10^4$ *yuan* |
| | | Robot Installation Density | % |
| | | AI-related company | |
| Digital economy index | | Internet users | $10^4$ |
| | | Internet Penetration Rate | % |
| | | Internet-related output | $10^4$ *yuan* |
| | | Internet-related employees | $10^4$ |
| | | Digital inclusive finance index | |
| AI technology level | | AI-pilot city | {0,1} |
| | | Intelligent construction city | {0,1} |
| | | AI-related output | $10^4$ *yuan* |
| | | AI infrastructure invest | $10^4$ *yuan* |
| | | AI infrastructure output | $10^4$ *yuan* |
| | | Intelligent manufacturing proportion | % |
| | | AI enterprises | |
| | | AI patents | |
| Carbon emissions | *ton* | Transportation, Construction, Industry | *ton* |
| | | Agriculture, Forestry, Land Use | *ton* |
| | | Waste, Electricity, Heating, Cooling | *ton* |

MAE, in contrast, calculates the average of the absolute differences between predicted and actual values. Unlike MSE, it treats all errors equally and is not influenced by outliers. MAE is often appreciated for its interpretability, as it retains the original unit of measurement of the target variable. Lower values of MAE also reflect better model performance, especially in terms of average deviation. The expression is given by:

$$MAE = \frac{1}{n}\sum_{i=1}^{n}|y_i - \hat{y}_i| \tag{19}$$

$R^2$ quantifies the proportion of variance in the dependent variable that is explained by the model. It serves as a goodness-of-fit metric, with values closer to 1 indicating a higher explanatory power. It is computed as:

$$R^2 = 1 - \frac{\sum_{i=1}^{n}(y_i - \hat{y}_i)^2}{\sum_{i=1}^{n}(y_i - \hat{y})^2} \tag{15}$$

where $\hat{y}$ is the mean of observed values. A higher $R^2$ suggests that the model more effectively captures the variance in the target variable.

## 5.2 Model performance

A series of benchmark models is employed for comparative analysis, including classical statistical approaches and widely used machine learning and deep learning algorithms in Table 2. Specifically, the baseline methods encompass Linear Regression (LR), K-Nearest Neighbors (KNN), and Support Vector Machines (SVM), representing conventional regression and distance-based learners. In ensemble learning, Random Forest (RF) and Extreme Gradient Boosting (XGBoost) are selected due to their proven effectiveness in handling high-dimensional and non-linear datasets. In addition, several deep learning architectures are considered, including Deep Neural Network (DNN), Convolutional Neural Network (CNN), Recurrent Neural Network (RNN), and Long Short-Term Memory (LSTM) networks, which are commonly used for capturing complex temporal and spatial patterns. Furthermore, Hybrid Neural Network (HNN) and Deep & Cross Network (DCN) are included to examine the influence of cross-feature interactions and hybrid representations on predictive accuracy. By benchmarking MADCN against this diverse set of models, the study aims to demonstrate its superiority in modeling both the individual and interactive effects of multi-dimensional urban indicators on carbon emissions while enhancing prediction robustness and interpretability.

**Table 2**

Model predictive performance

| Model | Train set | | | Test set | | |
|---|---|---|---|---|---|---|
| | MSE | MAE | $R^2$ | MSE | MAE | $R^2$ |
| LR | 29011606.433 | 3135.936 | 0.596 | 29857011.691 | 3215.522 | 0.588 |
| KNN | 14949845.850 | 1041.458 | 0.792 | 15520953.022 | 1061.386 | 0.778 |
| SVM | 8385937.376 | 823.474 | 0.886 | 8463628.614 | 835.585 | 0.875 |
| RF | 7189042.309 | 1178.761 | 0.903 | 7276438.156 | 1287.863 | 0.891 |
| XGBoost | 4367644.073 | 1074.732 | 0.937 | 4628550.118 | 1091.425 | 0.935 |
| DNN | 6463055.872 | 1278.255 | 0.913 | 6522627.962 | 1304.411 | 0.904 |
| CNN | 4422429.423 | 1203.891 | 0.938 | 4604832.665 | 1230.698 | 0.934 |
| LSTM | 85735871.703 | 4136.614 | 0.249 | 113656984.280 | 4799.3476 | 0.254 |
| RNN | 85726360.914 | 4136.054 | 0.250 | 113643198.503 | 4798.3455 | 0.256 |
| HNN | 1290975.628 | 714.873 | 0.951 | 26054673.485 | 1261.071 | 0.713 |
| DCN | 1614711.820 | 793.140 | 0.972 | 1652482.079 | 862.079 | 0.967 |
| **MADCN** | **395224.995** | **505.930** | **0.994** | **406151.063** | **612.304** | **0.991** |

LR serves as a baseline and exhibits the poorest performance among all models. With MSE values of 29,011,606.433 and 29,857,011.691 on the training and testing sets respectively, and $R^2$ scores of only 0.596 and 0.588, LR demonstrates its limitations in modeling non-linear and complex interactions within the data. The high MAE values (3135.936 and 3215.522) further indicate that linear assumptions fail to capture the true structure of the data, leading to substantial prediction errors. KNN provides an appreciable improvement over LR, reducing the MSE by nearly 50% on both sets. The $R^2$ scores improve to 0.792 on the training set and 0.778 on the testing set. However, KNN's performance is still not optimal, reflecting its sensitivity to the curse of dimensionality and its inability to generalize well when the data distribution is complex.

SVM emerges as a stronger candidate among traditional models. With an $R^2$ of 0.886 for training and 0.875 for testing, SVM effectively models the complex boundary structures in the data, suggesting its advantage in high-dimensional spaces. The relatively low MAE values (~823 on training and ~835 on testing) confirm that SVM can achieve tighter predictions around actual values. Nonetheless, there remains a slight performance gap when compared with ensemble methods. RF furthers this trend, achieving $R^2$ values of 0.903 (train) and 0.891 (test). RF significantly reduces the variance component of the prediction error through bagging, although it can still suffer from limited interpretability. The marginal overfitting observed (slightly higher $R^2$ on training) is typical for RF models and considered acceptable given the substantial improvement in overall predictive accuracy. XGBoost surpasses RF in MSE and MAE, achieving an $R^2$ of 0.937 on training and 0.935 on testing. These results reflect XGBoost's powerful gradient boosting strategy, which iteratively refines model residuals, making it highly efficient for modeling structured tabular data with complex feature interactions. The low discrepancy between training and testing $R^2$ indicates minimal overfitting, demonstrating strong generalization capabilities.

DNN delivers reasonable performance with $R^2$ values of 0.913 (train) and 0.904 (test). Although these are competitive figures, the higher MAE (~1278 and 1304) compared to XGBoost suggests that DNN struggles slightly more with capturing precise small-scale variations in the data. CNN outperforms DNN, achieving $R^2$ scores of 0.938 and 0.934. This is particularly noteworthy, as CNNs are typically associated with spatial data rather than tabular data; here, their success may be attributed to the model's ability to extract local patterns among feature groups. However, RNN and LSTM models perform poorly. Both models report extremely high MSEs (above 85 million for LSTM and RNN) and very low $R^2$ values (~0.25), indicating their inadequacy for this type of task. LSTM and RNN are designed to capture temporal dependencies, which are likely weak or non-existent in the present static feature set. Consequently, these models fail to model the underlying data structure appropriately, leading to severe underfitting. Among hybrid and more advanced models, HNN achieves an $R^2$ of 0.951 on the training set, suggesting excellent fitting ability. However, its testing $R^2$ drops to 0.713, signaling clear overfitting. The model captures noise in the training data, thus compromising its ability to generalize to unseen data. DCN demonstrates a remarkable balance between fitness and generalization. Achieving $R^2$ values of 0.972 and 0.967 for training and testing, respectively, the DCN model significantly reduces both MSE and MAE compared to previous models. The architecture's ability to explicitly model feature interactions via cross terms, coupled with the depth of a standard feedforward network, allows it to capture both low-order and high-order patterns effectively.

MADCN delivers the best predictive performance among all evaluated models. Concerns about overfitting are mitigated by rigorous validation. Training ($R^2$=0.994, MSE=395,224.995) and test ($R^2$=0.991, MSE=406,151.063) metrics align closely, indicating generalization. MADCN's architecture (multi-head attention, cross network) prioritizes meaningful features and stable interactions, reducing noise sensitivity. The 275-city dataset (2009–2021) covers diverse regions, stages, and scales, ensuring robustness. Superior performance confirms reliability. With $R^2$ values of 0.994 for the training set and 0.991 for the testing set, and extremely low MSE values (395,224.995 and 406,151.063), MADCN almost perfectly models the carbon emission outcomes. The improvement likely stems from the incorporation of multiple attention mechanisms that dynamically weigh feature importance and interaction strength, allowing the model to focus on the most influential predictors in different contexts. The comparative analysis demonstrates that conventional linear models are insufficient for the carbon emission dataset, while ensemble-based and deep learning methods significantly enhance predictive accuracy. Advanced architectures such as DCN and especially MADCN offer superior performance by capturing complex feature relationships, demonstrating high fitting capability and excellent generalization ability. These findings underscore the importance of model selection tailored to data complexity, and they affirm the value of hybrid neural architectures in carbon emission prediction tasks.

### 5.3 Model interpretation results

**Fig. 5** presents the SHAP force plots for two individual samples provide important insights into the feature contributions toward the predicted carbon emissions within the MADCN model. In the first sample, features such as AI Technology Level, NQPF, and Population are shown to significantly push the prediction to a higher value compared to the model's base value, as indicated by the strong red segments. Among them, AI Technology Level exerts the most substantial positive influence, suggesting that for this sample, a higher AI technological advancement level is associated with increased carbon emissions. Similarly, NQPF, which typically represents emerging industries and innovative economic activities, also contributes to an elevation in emissions in this context, possibly reflecting the energy-intensive nature of the early development stages of such sectors. The Population factor also has a notable positive effect, aligning with the common understanding that greater urban populations typically correspond to higher total emissions.

Conversely, in the second sample, although AI Technology Level remains a positive driver (red contribution), its relative magnitude is smaller compared to the first sample. Interestingly, features like City Size, GDP, and Population exhibit strong blue contributions, meaning they pull the prediction downward, thereby reducing the expected carbon emissions. This outcome suggests that in this specific context, larger cities and higher economic outputs are associated with better energy efficiencies or perhaps greater implementation of low-carbon technologies, offsetting the anticipated rise in emissions.

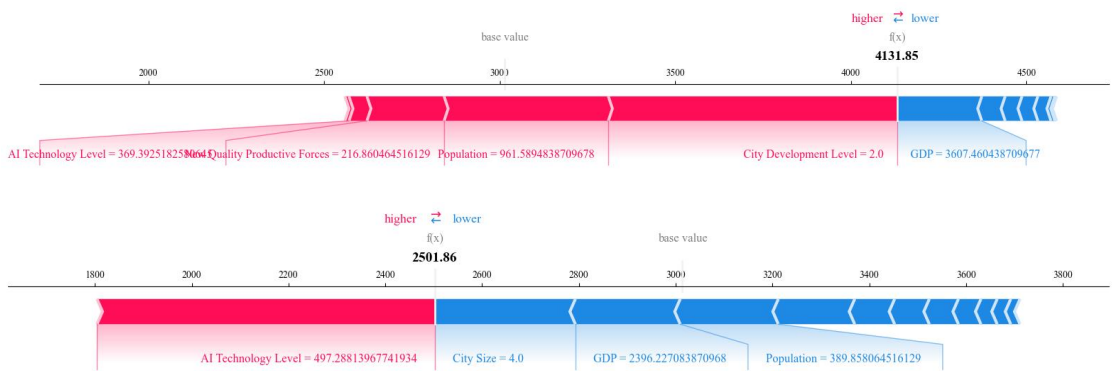

**Figure. 5.** Random single sample

**Fig. 6(a)** presents a feature importance bar plot based on SHAP values, which illustrates the average magnitude of feature contributions to the model's predictions of carbon emissions. The most prominent variables are Population, City Size, and Urbanization Rate, consistent with conventional understanding that demographic scale and urban morphology are dominant drivers of energy consumption and emission patterns. NQPF, Digital Economy Index, and AI Technology Level exhibit different levels of importance, each providing unique insights into the evolving dynamics of low-carbon development. NQPF rank relatively low in average SHAP importance. Despite their strategic role in fostering sustainable industries, their modest contribution suggests that the carbon-reduction potential of advanced manufacturing and emerging industries has not yet been fully realized. It is plausible that NQPF are still in an incubation phase, wherein technological advancement has not yet significantly shifted energy consumption structures or efficiency levels.

Similarly, Digital Economy Index shows a very low importance and a highly concentrated distribution of impact values. This outcome indicates that, at least in the short term, digital transformation initiatives may not exert a strong direct influence on carbon emissions. Digitalization processes often lead to rebound effects or structural inertia that dilute immediate environmental benefits, particularly when infrastructure or policy frameworks lag behind. In contrast, AI Technology Level holds a middle-upper rank among the features. More importantly, subsequent analyses reveal that higher AI technology levels are associated with notable negative SHAP values, implying significant carbon reduction contributions. This aligns with emerging empirical evidence that artificial intelligence enhances operational efficiencies, optimizes energy systems, and facilitates smart governance, thereby driving measurable emission reductions even at early stages of deployment.

These findings suggest that while technological and economic innovations are critical for long-term decarbonization, their short-term effects vary substantially. A more nuanced modeling of indirect and lagged mechanisms is warranted to better capture the transformative impact of new productive forces and digital economies in future studies.

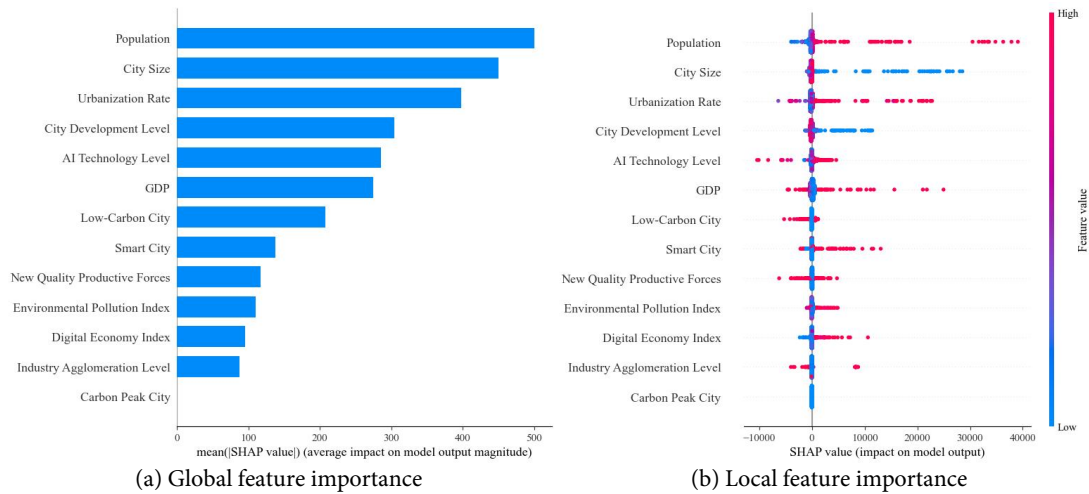

(a) Global feature importance         (b) Local feature importance

**Figure. 6.** Variable feature importance

**Fig. 6(b)** illustrates the distribution of individual sample contributions across different features. Each point represents a sample-specific SHAP value, with its color indicating the magnitude of the original feature value. This visualization allows for a granular understanding of both the magnitude and directionality of feature influences on carbon emission predictions. Focusing on NQPF, the plot shows a tight concentration of SHAP values near zero, with a slight rightward skew for higher feature values. This suggests that while higher levels of new productive forces exert a weak positive impact on predicted carbon emissions, their overall influence remains minimal. The pattern implies that despite policy emphasis on industrial upgrading, NQPF have not yet exerted a substantial decarbonization effect. Possible reasons include the current high energy demands of emerging industries and the time lag inherent in structural transitions towards low-carbon production.

The Digital Economy Index exhibits an even more concentrated distribution, centered tightly around zero, with almost no discernible pattern between high and low feature values. This observation confirms that, within the present temporal scope, the development of the digital economy has a negligible direct impact on carbon emissions. The short-term influence remains limited, possibly due to the dual effects of energy-intensive data infrastructure offsetting gains from digital efficiencies. Conversely, AI Technology Level presents a markedly different pattern. Samples with higher AI technology values (represented in red) predominantly shift towards negative SHAP values. This indicates that increased adoption of AI technologies correlates with lower predicted carbon emissions, substantiating the theory that AI facilitates energy optimization, predictive maintenance, and adaptive resource management. The clear negative association demonstrates that AI's contribution to emission mitigation is already becoming operationally significant.

### 5.4 Variable interaction effect

**Fig. 7(a)** illustrates the interaction between NQPF and the Digital Economy Index in terms of their SHAP value impact on carbon emissions. Overall, when the NQPF remain at low levels (below 100), their influence on carbon emissions is relatively minor, as reflected by the low and concentrated SHAP values. As the NQPF increase beyond 200, the distribution of SHAP values becomes more dispersed, indicating heterogeneous effects across samples. It is notable that across different levels of Digital Economy Index—ranging from low (blue) to high (red)—the distribution pattern does not significantly change, suggesting that, at the current stage, the digital economy exerts limited modulation on the carbon emission effects of new productive forces. This phenomenon can be attributed to the relatively immature stage of NQPF, where innovations have not yet been fully deployed to realize their carbon-reduction potential. Despite high levels of NQPF, the green transition benefits are not yet significant, possibly due to the lag in industrial application and insufficient ecosystem maturity. Moreover, the Digital Economy Index appears to have minimal short-term amplifying or mitigating effects, supporting the interpretation that digital economy developments are still emerging and have yet to exert substantial environmental benefits in the current timeframe.

**Fig. 7(b)** presents the interaction effect between NQPF and AI Technology Level on carbon emissions. When NQPF are low, the SHAP values remain consistently small across varying AI technology levels. However, as NQPF increase beyond 200, a clear trend emerges: higher AI Technology Level (denoted by red coloring) is associated with more negative SHAP values, indicating a stronger carbon reduction effect. This suggests a synergistic relationship wherein the positive impacts of NQPF are significantly enhanced when complemented by advanced AI technologies. This result confirms that AI technologies have already begun to exhibit a tangible contribution toward carbon mitigation, particularly in contexts where advanced productive forces are available. The underlying mechanism likely relates to AI's role in optimizing production processes, reducing resource wastage, and enhancing energy efficiency through intelligent decision-making and automation. Therefore, fostering high levels of AI technology development alongside the advancement of NQPF is critical for maximizing their green transition potential and realizing meaningful reductions in carbon emissions.

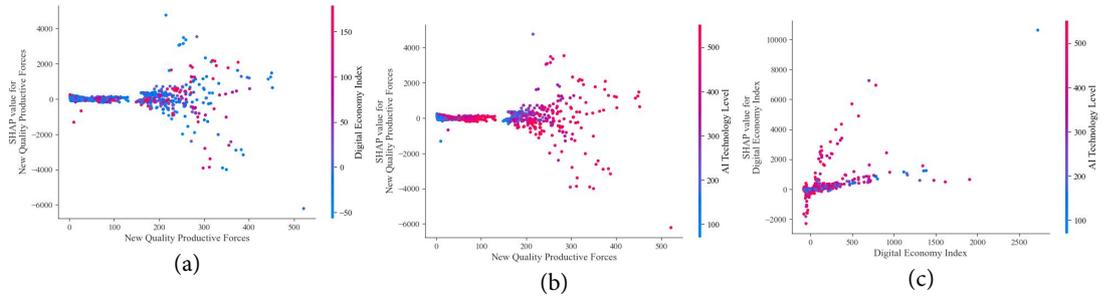

**Figure. 7.** Interaction effect of NQPF, digital economy, and AI

**Fig. 7(c)** analyzes the interaction between the Digital Economy Index and the AI Technology Level. In the range where the Digital Economy Index remains low (under 500), the SHAP values are densely clustered near zero, implying limited impact on carbon emissions regardless of the AI development level. It is only when the Digital Economy Index reaches higher values that the SHAP values display greater dispersion. Particularly, with higher AI Technology Level (indicated by red hues), SHAP values tend to become negative, signifying a more substantial carbon reduction effect. This observation indicates that the Digital Economy Index alone, at its current development level, has not yet achieved significant environmental gains. However, when integrated with advanced AI technologies, the digital economy begins to drive noticeable carbon emission reductions. In essence, the environmental benefits of the digital economy are contingent upon its deep integration with AI-driven technologies, such as intelligent logistics, smart energy management, and predictive maintenance systems. In the short term, the digital economy remains at a relatively nascent stage, resulting in limited direct effects on carbon emissions. Nevertheless, its long-term potential, especially when coupled with AI advancements, could become a major force in promoting sustainable and low-carbon economic transitions.

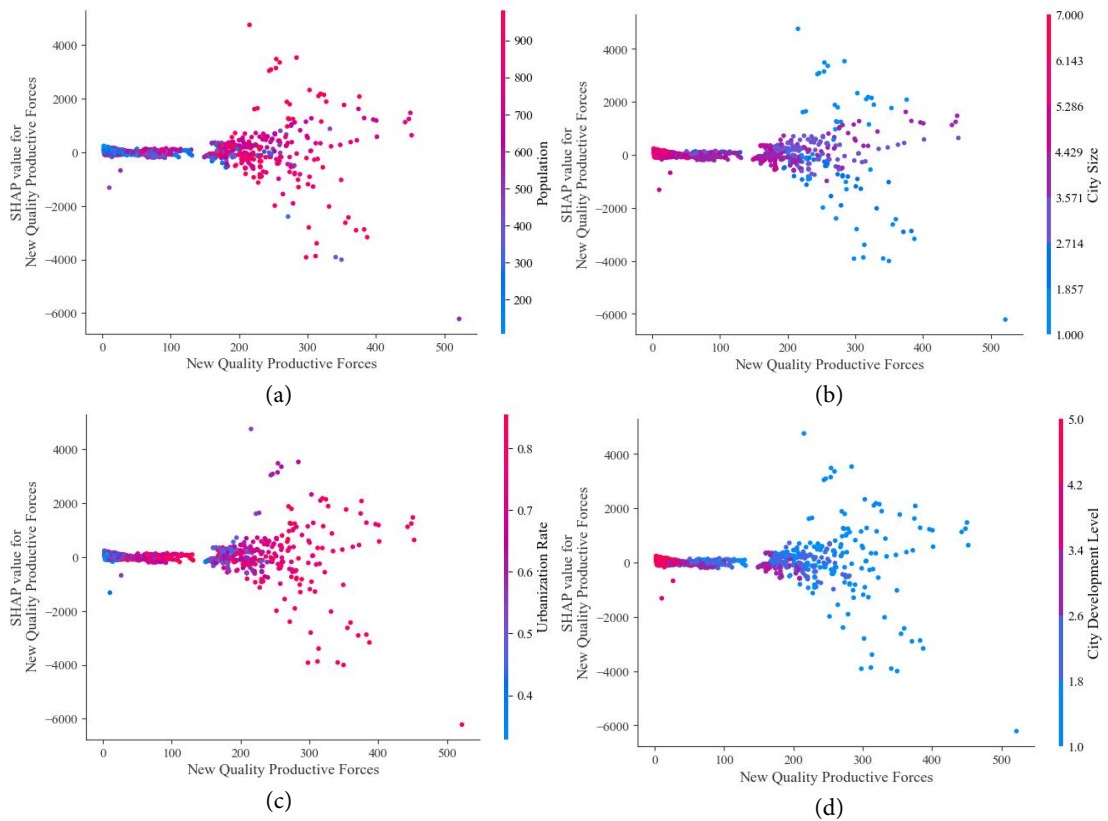

**Figure. 8.** Interaction effect between NQPF and top driving factors

**Fig. 8(a)** examines the relationship between NQPF and Population in shaping carbon emissions. The majority of the data points exhibit a positive SHAP value with increasing Population, particularly when NQPF are relatively higher (>200). This indicates that in regions with higher levels of new productive forces, a larger population tends to amplify carbon emissions rather than mitigate them. In contrast, when NQPF are low (<100), population size appears to exert a minimal or even slightly negative impact. This suggests that underdeveloped productive systems are insufficient to offset the carbon footprint associated with population growth. A possible explanation lies in the fact that regions with higher populations and emerging productive forces may not have fully transitioned into green economic structures, thus larger populations still correspond to higher energy demands and subsequent emissions. Moreover, the distribution of colors reveals that larger population sizes (depicted by darker red tones) are associated with a wider spread of positive SHAP values, reinforcing the notion that demographic pressure can dilute the potential emission-reduction effects of technological progress if not accompanied by corresponding green transformations.

**Fig. 8(b)** captures the interplay between NQPF and City Size. Here, we observe that larger cities (indicated by the darker colors) tend to have slightly more positive SHAP values at moderate levels of NQPF (approximately 150–300). This implies that urban agglomerations could enhance carbon emissions even in the presence of advancements in productive capabilities. However, at very high levels of NQPF (>400), the impact becomes more neutral or even slightly negative, suggesting that once a city achieves sufficient technological modernization, its size no longer amplifies emissions to the same extent. One possible reason is that large cities, while initially contributing to increased emissions due to concentrated industrial activity and transportation demands, may eventually leverage advanced productive forces to implement more efficient energy systems, better public transportation, and stricter environmental regulations. Nonetheless, the spread and variability of SHAP values at moderate city sizes indicate a heterogeneous impact, pointing to differing city-level strategies and capacities in integrating new technologies into sustainable urban development.

**Fig. 8(c)** shows how Urbanization Rate moderates the influence of NQPF on carbon emissions. A notable pattern is observed where higher Urbanization Rates (represented by darker red shades) tend to shift the SHAP values upward as NQPF increase. Particularly in regions with an Urbanization Rate above 0.7, the SHAP values for carbon emissions tend to be positive when NQPF are between 150 and 300. This suggests that urbanization, when not accompanied by sustainable infrastructure and green technologies, can exacerbate carbon emissions despite improvements in productive capabilities. In early stages of technological adoption, rapid urban growth often demands significant energy consumption for transportation, construction, and industrial operations, thereby overshadowing any emission-reduction benefits from new technologies. This finding highlights that merely increasing NQPF is insufficient; synchronized advancements in urban planning, energy systems, and public services are necessary to harness urbanization for decarbonization effectively. Without such integration, higher Urbanization Rates can negate the potential benefits of emerging productive forces.

**Fig. 8(d)** focuses on the interaction between NQPF and City Development Level. Interestingly, a higher City Development Level (depicted by redder points) is associated with slightly lower SHAP values, especially as NQPF increase beyond 200. This suggests that in cities with more advanced development stages, the positive impact of NQPF on emissions tends to be mitigated. In other words, better-developed cities are more capable of utilizing new productive capabilities to control or even reduce carbon emissions. This pattern supports the notion that maturity in infrastructure, governance, and innovation ecosystems enables cities to transform technological advancements into tangible environmental benefits. Well-developed cities are likely to have better enforcement of environmental policies, higher public awareness, and more comprehensive integration of clean energy technologies. However, it is important to note that at lower City Development Levels, the SHAP values tend to be more positive, implying that less developed cities might initially struggle to convert new productive forces into emission reductions, possibly due to limited absorptive capacities or systemic inefficiencies.

## 6. Conclusion and policy implications

This study investigates the complex relationships between new quality productive forces (NQPF), digital economy development, artificial intelligence (AI) technological advancement, and urban carbon emissions in China. By constructing a novel Multi-head Attention Deep & Cross Network (MADCN) framework, the research addresses the limitations of conventional machine learning and deep learning models, particularly their inability to capture feature interactions and provide interpretability. Using a comprehensive panel dataset of 275 Chinese cities from 2009 to 2021, the study demonstrates that the MADCN model significantly outperforms baseline models in predictive accuracy, achieving an $R^2$ of 0.991 on the test set. Moreover, SHAP analysis reveals that while traditional factors such as population size, city size, and urbanization rate remain dominant drivers of carbon emissions, emerging variables—specifically NQPF, the Digital Economy Index, and AI Technology Level—exhibit differentiated impacts. NQPF currently has a weak and slightly positive influence on emissions, suggesting that their carbon mitigation potential is still latent. The Digital Economy Index also exhibits limited short-term impact, with SHAP values tightly concentrated around zero. In contrast, AI Technology Level displays a significant negative association with carbon emissions, indicating that AI adoption is already delivering tangible decarbonization benefits. Interaction analyses further reveal that the emission-reduction potential of NQPF and digital economy initiatives is substantially amplified when coupled with high levels of AI development, emphasizing the importance of technological integration for achieving sustainable urban transitions.

This study advances theoretical understanding in two key ways. First, integrating NQPF provides a novel theoretical pathway to capture China's unique urban green transition dynamics. Unlike fragmented analyses of isolated factors (e.g., digitalization or AI), NQPF operationalizes the systemic fusion of technological innovation, industrial upgrading, and institutional design—revealing that green transformation in Chinese cities is driven by synergies between these elements rather than single-factor advancement. SHAP interaction results confirm that NQPF amplifies the decarbonization effects of digital and AI technologies, offering a new framework to model how policy-driven productive forces reshape environmental outcomes. Second, the model's design—leveraging multi-head attention to capture contextual heterogeneity and cross networks to quantify feature interactions—transcends China's context. For other countries, NQPF can be adapted to region-specific constructs (e.g., "green industrial policy bundles" in Europe or "technology-ecosystem synergies" in Southeast Asia), while retaining its core capability to unpack complex driver interactions, addressing a critical limitation of existing models that often fail to balance context specificity with systemic analysis.

Based on observed urban heterogeneities, targeted policy recommendations are proposed. Regionally, eastern cities (with mature digital/AI infrastructure) should prioritize AI-driven optimization of industrial energy use and smart urban planning to capitalize on NQPF-AI synergies, while western cities (with emerging digital ecosystems) need foundational investments in digital infrastructure and AI talent cultivation to unlock NQPF's latent potential. In terms of city size, megacities, facing concentrated emissions from transportation and construction, should deploy AI-integrated systems for traffic management and green building oversight, whereas smaller cities, where industrial emissions dominate, should focus on NQPF-led industrial upgrading to avoid carbon-intensive growth paths. Additionally, all cities must align digital economy policies with AI innovation, mandating AI-enabled energy management in data centers and incentivizing smart logistics—to amplify decarbonization effects.

## Date Availability Statement

All data are included in this published article and its supplementary information.

## Funding

This work was supported by the National Social Science Fund of China (No. 23FZXB028).